\documentclass[letterpaper, 10 pt, conference]{ieeeconf}  

\IEEEoverridecommandlockouts                              

\usepackage{algorithm}
\usepackage{algorithmic}
\usepackage[font=footnotesize]{caption}
\usepackage{graphicx}
\usepackage{xcolor}
\let\labelindent\relax
\usepackage{enumitem}
\usepackage{url}

\usepackage{tikz}
\newcommand*\circled[1]{\tikz[baseline=(char.base)]{
            \node[shape=circle,draw, color=black,inner sep=0.8pt, minimum size=2pt] (char) {#1};}}
            
\newcommand{\rpoint}[1]{\circled{{\fontfamily{pcr}\selectfont\footnotesize{#1}}}}

\usepackage{dblfloatfix}

\usepackage{titlesec}

\title{\LARGE \bf
R-SNN: An Analysis and Design Methodology for Robustifying Spiking Neural Networks against Adversarial Attacks through Noise Filters for Dynamic Vision Sensors\\
}

\author{Alberto Marchisio\textsuperscript{1,*}\thanks{*These authors contributed equally to this work.}, Giacomo Pira\textsuperscript{2,*}, Maurizio Martina\textsuperscript{2}, Guido Masera\textsuperscript{2}, Muhammad Shafique\textsuperscript{3}\\
\textit{\textsuperscript{1}Institute of Computer Engineering, Technische Universit{\"a}t Wien, Vienna, Austria}\\
\textit{\textsuperscript{2}Department of Electronics and Telecommunication, Politecnico di Torino, Turin, Italy}\\
\textit{\textsuperscript{3}Division of Engineering, New York University, Abu Dhabi, UAE}\\ 
\textit{Email: alberto.marchisio@tuwien.ac.at, giacomo.pira@studenti.polito.it}\\
\textit{\{maurizio.martina, guido.masera\}@polito.it, muhammad.shafique@nyu.edu}\\
\vspace*{-10pt}}

\usepackage{fancyhdr,lipsum}
\fancypagestyle{firstpage}{
  \fancyhf{}
  \fancyhead[C]{To appear at the 2021 IEEE/RSJ International Conference on Intelligent Robots and Systems (IROS 2021).}
  \fancyfoot[C]{\thepage}
}

\begin{document}

\maketitle
\thispagestyle{empty}
\pagestyle{empty}
\thispagestyle{firstpage}






\begin{abstract}

Spiking Neural Networks (SNNs) aim at providing energy-efficient learning capabilities 
when implemented on neuromorphic chips with event-based Dynamic Vision Sensors (DVS). This paper studies the robustness of SNNs against adversarial attacks on such DVS-based systems, and proposes \textit{\mbox{R-SNN}}, a novel methodology for robustifying SNNs through efficient DVS-noise filtering. We are the first to generate adversarial attacks on DVS signals (i.e., frames of events in the spatio-temporal domain) and to apply noise filters for DVS sensors in the quest for defending against adversarial attacks. Our results show that the noise filters effectively prevent the SNNs from being fooled. The SNNs in our experiments provide more than 90\% accuracy on the DVS-Gesture and NMNIST datasets under different adversarial threat models.

\normalfont{\textit{Index Terms: Spiking Neural Networks, SNNs, Deep Learning, Adversarial Attacks, Security, Robustness, Defense, Filter, Perturbation, Noise, Dynamic Vision Sensors, DVS, Neuromorphic, Event-Based, DVS-Gesture, NMNIST.}}
\end{abstract}

\section{Introduction}

Spiking Neural Networks (SNNs) aim at providing energy-efficient learning capabilities in a wide variety of machine learning applications, e.g., autonomous driving~\cite{Zhou2020SNNAD}, healthcare~\cite{Gonzalez2020SNNhealthcare}, and robotics~\cite{Tang2018SNNrobotics}. Unlike traditional (i.e., non-spiking) Deep Neural Networks (DNNs), the SNNs are biologically plausible, enabling event-based communication between neurons which simulate the human brain's processing in a relatively closer manner~\cite{Kasinski2011IntroSNN}. Moreover, the results both in terms of power/energy efficiency and real-time classification performance make the SNNs appealing for being implemented in resource-constrained embedded systems~\cite{Capra2020SurveyDNN}. By leveraging the spike-based communication between neurons, SNNs exhibit a lower computational load, as well as a reduction in the latency, compared to the equivalent DNN implementations~\cite{Deng2020ComparisonANNSNN}. 

Along with the development of efficient SNNs implemented on specialized neuromorphic accelerators (e.g., IBM TrueNorth~\cite{Merolla2014Truenorth} and Intel Loihi~\cite{Davies2018Loihi}), another advancement in the field of neuromorphic hardware has come from the new generation of the Dynamic Vision Sensor (DVS), i.e., an event-based camera sensor~\cite{Lichtsteiner2006DVS}. Unlike a classical frame-based camera, the DVS emulates the behavior of the human retina, by recording the information in form of a sequence of spikes, which are generated every time a change of light intensity is detected. 
The event-based behavior of these sensors pairs well with SNNs implemented onto the neuromorphic hardware, i.e., the output of a DVS camera can be used as the direct input of an SNN to elaborate events in real-time.

\subsection{Target Research Problem and Scientific Challenges}

Similar to the case of traditional DNNs, the trustworthiness of SNNs is also threatened by adversarial attacks, i.e., small and imperceptible input perturbations aiming at crafting the network's correct functionality. Although some preliminary studies have been conducted~\cite{Bagheri2018AdvTrainingSNN}\cite{Marchisio2019SNNAttack}\cite{Sharmin2019ACA}\cite{Liang2020ExploringAA}, such a problem is relatively new and unexplored for practical SNN-based systems. In particular, DVS-based systems have not been investigated for SNN security. As a starting point, the methods for designing robust SNNs can be derived from the recent advancements of the defense mechanisms for DNNs, where studies have focused on adversarial learning algorithms~\cite{Madry2017TowardsDL}, loss/regularization functions~\cite{Zhang2019TradeoffRobustnessAccuracy}, and image preprocessing~\cite{Khalid2019FAdeML}. 
The latter approach basically consists of suppressing the adversarial perturbation through dedicated filtering. Noteworthy, for the SNN-based systems fed by DVS signals, the attacks and preprocessing-based defense techniques for frame-based sensors cannot be directly applied due to differences in the signal properties. Therefore, specialized noise filters for DVS sensors~\cite{LinaresBarranco2019FilterDVS} must be employed.

As per our knowledge, the impact of filtering on DVS sensors for secure neuromorphic computing is an unexplored and open research problem. Towards this, we devise \textit{R-SNN}, a novel methodology employing attack-resistant noise filters on DVS signals as a defense mechanism for robustifying SNNs against adversarial attacks. Since the DVS cameras contain also the temporal information, the generation of adversarial perturbation is technically different w.r.t. traditional adversarial attacks on images, where only the spatial information is considered. Hence, the temporal information needs to be leveraged for developing a robust defense.


\subsection{Motivational Case Study}

\begin{figure}[h]
    \centering
    \includegraphics[width=\linewidth]{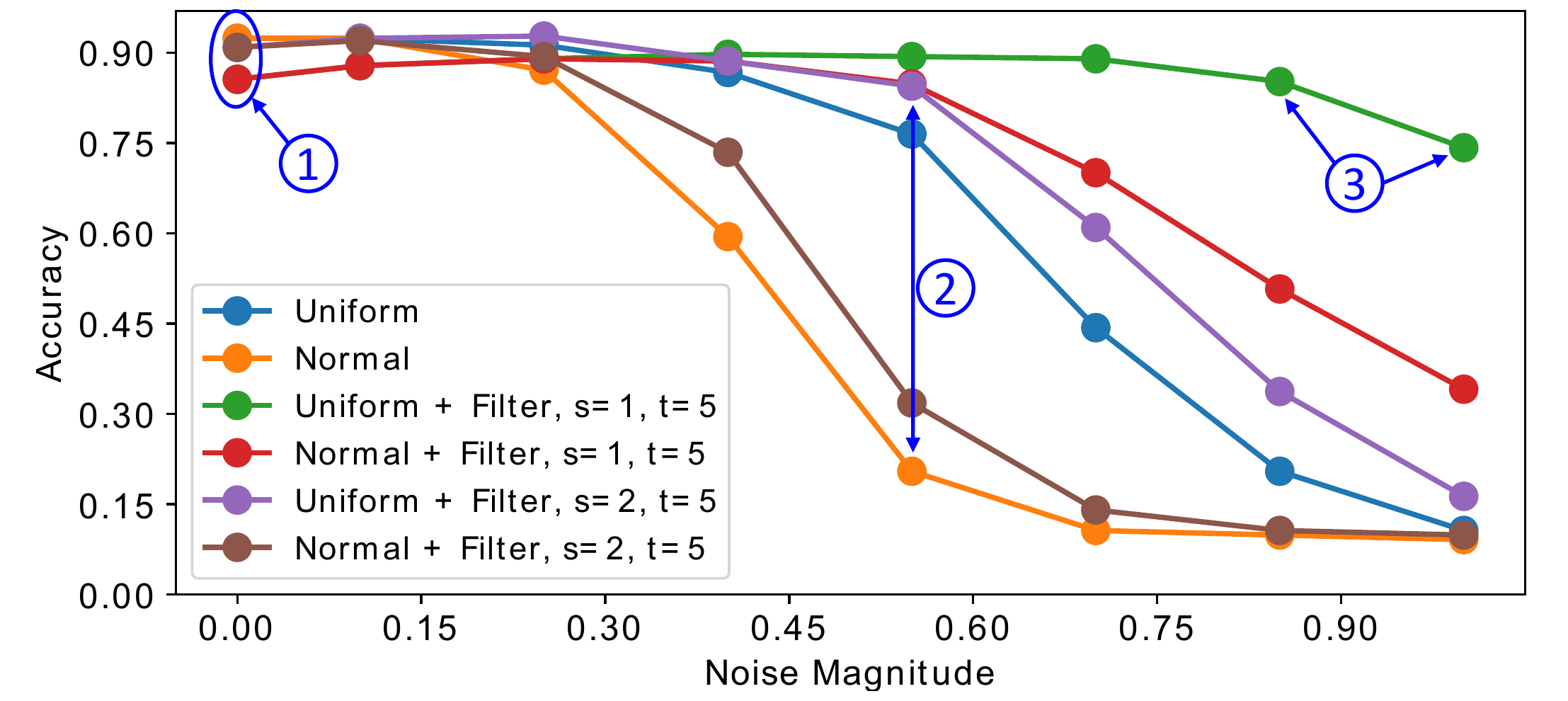}
    \caption{Analyzing the impact of applying the normal and uniform noise to the DVS-Gesture dataset.}
    \label{fig:noise}
    \vspace*{-10pt}
\end{figure}

As a preliminary study for motivating our research in the above-discussed directions, we perform the following experiments. We trained a 4-layer Spiking CNN, with 2 convolutional layers and 2 fully-connected layers, for the DVS-Gesture dataset~\cite{Amir2017DVSgesture} using the SLAYER method~\cite{Shrestha2018SLAYER}, using an ML-workstation with two Nvidia GeForce RTX 2080 Ti GPUs. For each frame of events, we perturb the testing dataset by injecting uniform and normally-distributed random noise and measure the classification accuracy. Moreover, to mitigate the effect of the perturbations, the filter of~\cite{LinaresBarranco2019FilterDVS} is applied, with different spatio-temporal parameters ($s$ and $t$). The accuracy results w.r.t. different noise magnitude are shown in Fig.~\ref{fig:noise}. As indicated by pointer~\rpoint{1} in Fig.~\ref{fig:noise}, the filter slightly reduces the accuracy of the SNN when no noise is applied. However, in the presence of noise, the SNN becomes much more robust when the filter is applied. For example, when considering normal noise with a magnitude of 0.55, the filter with $s=1$ and $t=5$ contributes to 64\% accuracy improvement; see pointer~\rpoint{2}. Such a filter works even better when uniformly-distributed noise is applied. Indeed, the perturbations with large magnitude of 0.85 and 1 are filtered out well, because the SNN maintains a relatively high accuracy of 85\% and 74\%, respectively; see pointer~\rpoint{3}.

\subsection{Our Novel Contributions}

\begin{figure}[h]
    \centering
    \vspace*{5pt}
    \includegraphics[width=\linewidth]{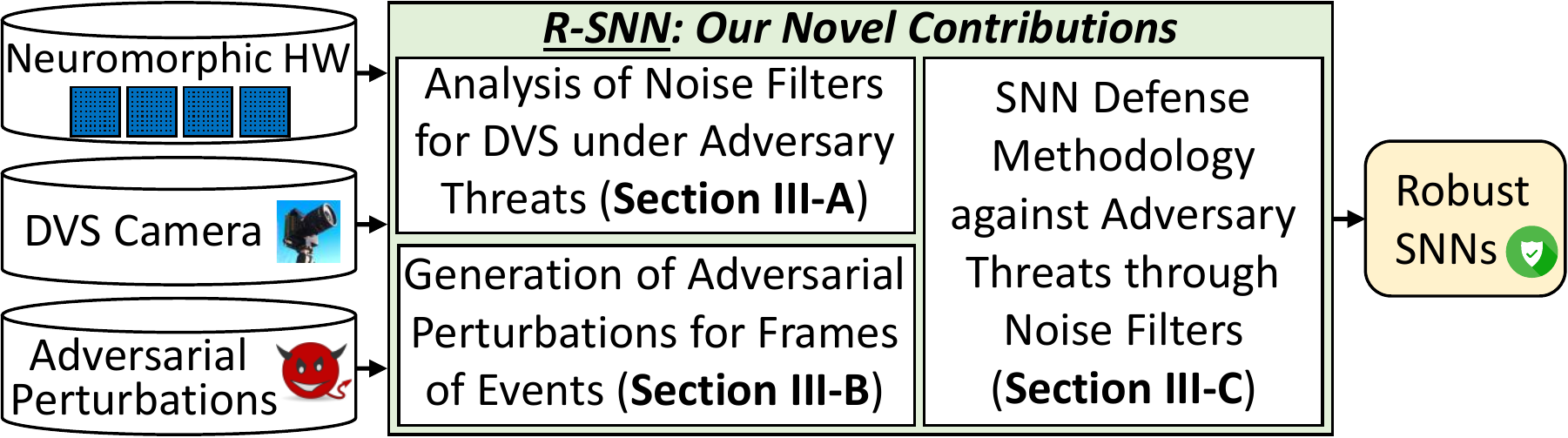}
    \caption{Overview of our novel contributions and methodology.}
    \label{fig:novel_contrib}
    \vspace*{-10pt}
\end{figure}

To address the above-discussed scientific problem, we propose \textit{R-SNN}, an analysis and design methodology for robustifying SNNs. Our \textbf{key contributions} are as follows (see Fig.~\ref{fig:novel_contrib}).

\begin{itemize}[leftmargin=*]
    \item We analyze the impact of noise filtering for DVS under multiple adversary threat models, i.e., by placing the filter at different stages of the system, or assuming different knowledge of the adversary. (\textbf{Section~\ref{subsec:threat_models}})
    \item We generate adversarial perturbations for the DVS signal to attack SNNs. (\textbf{Section~\ref{subsec:attack_DVS}})
    \item \textit{R-SNN Design Methodology:} we propose a methodology to apply specialized DVS-noise filters for increasing the robustness of SNNs against adversarial attacks. (\textbf{Section~\ref{subsec:methodology}})
    \item Our experimental results exhibit high SNN robustness against adversarial attacks, under different adversary threat models. (\textbf{Section~\ref{sec:results}})
    \item For reproducible research, we release the code of the \textit{R-SNN} filtering methodology for DVS-based SNNs on GitHub\footnote{\url{https://github.com/albertomarchisio/R-SNN}}.
\end{itemize}

\section{Background}

\subsection{Spiking Neural Networks (SNNs)}

SNNs, the third generation NNs~\cite{Maas1997ThirdGenerationSNN}, exhibit better biological plausibility compared to the traditional DNNs. Indeed, the event-based communication between neurons in SNNs resembles the human brain's functionality. Another key advantage of SNNs over the traditional DNNs is their improved energy-efficiency when implemented on Neuromorphic chips like Intel Loihi~\cite{Davies2018Loihi} or IBM TrueNorth~\cite{Merolla2014Truenorth}. Moreover, the recent development of DVS sensors~\cite{Lichtsteiner2006DVS} has further reduced the energy consumption of the complete system.

\begin{figure}[h]
	\centering
	\includegraphics[width=\linewidth]{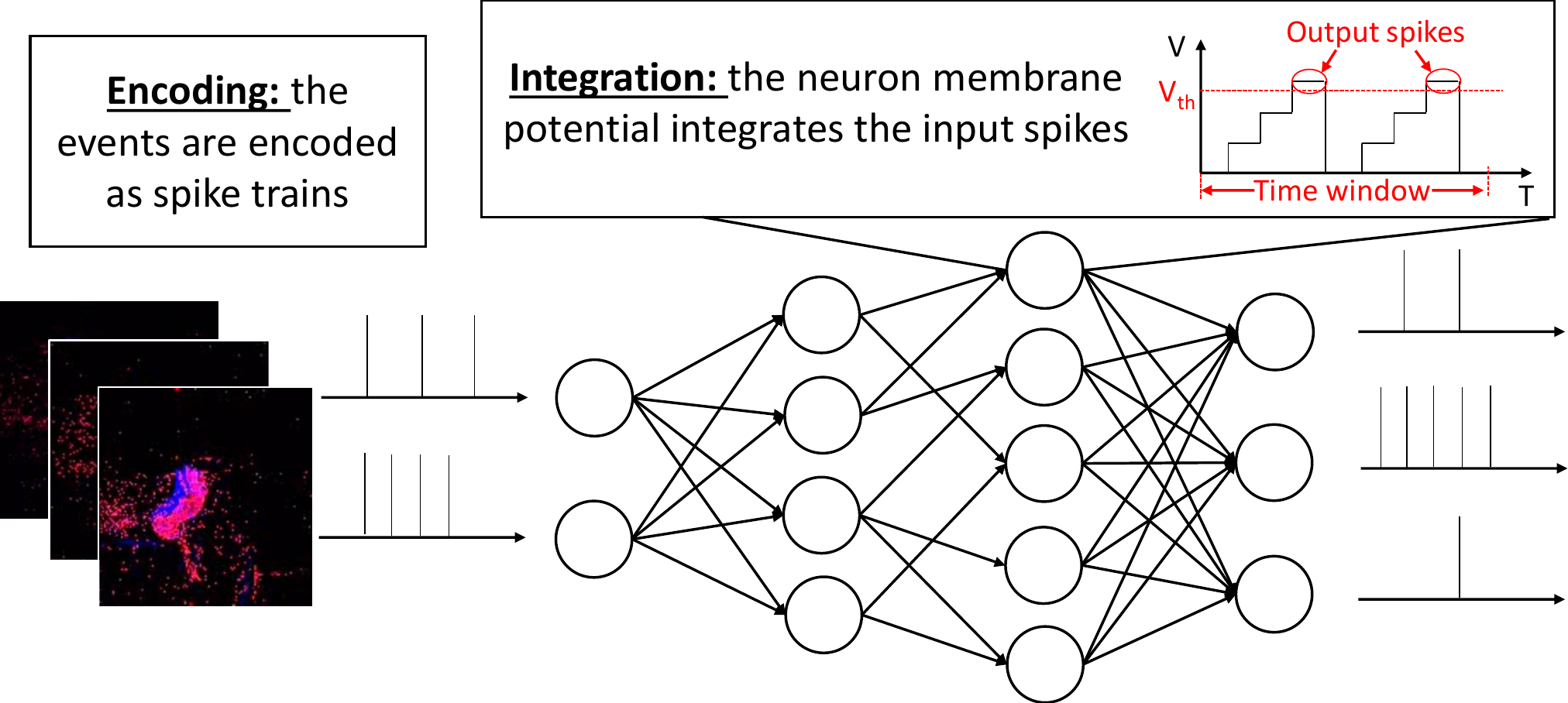}
	\caption{Overview of an SNN's functionality, focusing on the information encoding into spike trains and the integration of spikes into the membrane potential.}
	\label{fig:SNN}
\end{figure}

An example of the SNNs' functionality is shown in Fig.~\ref{fig:SNN}. The input is coded into spikes, which propagate to the output through the neurons' synapses. The most common encoding scheme is the rate encoding~\cite{Kasinski2011IntroSNN}, and the neurons integrate the incoming spikes to increase their membrane potential. Every time the potential overcomes a certain threshold, an output spike is emitted.

\subsection{Noise Filters for Dynamic Vision Sensors}

\textbf{Event-based cameras}~\cite{Lichtsteiner2006DVS} are bio-inspired sensors for the acquisition of visual information, directly related to the light variations in the scene. The DVS cameras work asynchronously, not recording frames with a precise timing. Instead, the sensors record negative and positive brightness variations in the scene. Thus, each pixel encodes a brightness change in the scene. Pixels are independent, and can record both positive and negative light variations. Compared to classical frame-based image sensors, the event-based sensors consume significantly less power, since the data is recorded only when a brightness variation is detected in the scene. This means that, in the absence of light changes, no information is recorded, leading close to zero power consumption. Hence, DVS sensors can be efficiently deployed at the edge and directly coupled to neuromorphic hardware for SNN-based applications.

\begin{figure}[h]
    \centering
    \includegraphics[width=.6\linewidth]{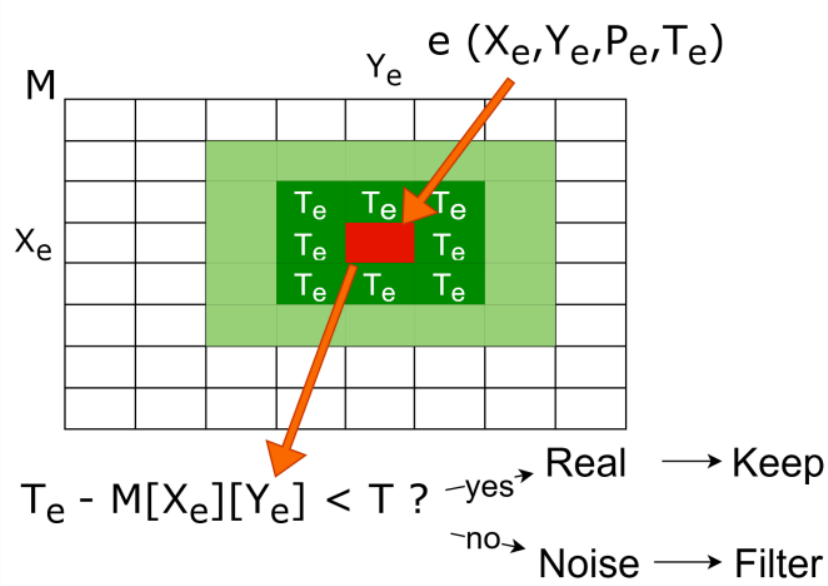}
    \caption{Functionality of the noise filter for frames of events.}
    \label{fig:filter}
    \vspace*{-10pt}
\end{figure}

DVS sensors are mainly affected by background activity noise, caused by thermal noise and junction leakage current~\cite{Nozaki2017ParasiticDVS}. When the DVS is stimulated, a neighborhood of pixels is usually active at the same time, generating events. Therefore, the real events show a higher spatio-temporal correlation than the noise-related events. This empirical observation is exploited for filtering out the noise~\cite{LinaresBarranco2019FilterDVS}. The events are associated with a spatio-temporal neighborhood, within which the correlation between them is calculated. If the correlation is lower than a certain threshold, the events are likely due to noise and thus are filtered out; otherwise they are kept. 
The procedure is reported in Algorithm~\ref{alg:filterDVS}, where $s$ and $t$ are the only parameters of the filter and are used to set the dimensions of the spatio-temporal neighborhood. The larger $s$ and $t$ are, the lower the number of events are filtered out. As shown in the example of Fig.~\ref{fig:filter}, the decision of the filter is made by the comparison between $t_e - M[x_e][y_e]$ and $t$ (lines 15-16 of Algorithm~\ref{alg:filterDVS}). If the first term is lower, then the event is filtered out. 

\begin{figure}[h]
\vspace*{-10pt}
\begin{algorithm}[H]
\caption{\textbf{:} Noise filter in the spatio-temporal domain.}
\label{alg:filterDVS}
\begin{small}
\begin{algorithmic}[1]
\STATE Being $E$ a list of events of the form $(x,y,p,t)$
\STATE Being $(x_e,y_e,p_e,t_e)$ the x-coordinate, the y-coordinate, the polarity and the timestamp of the event $e$ respectively
\STATE Being $M$ a $128 \times 128$ matrix
\STATE Being S and T the spatial and temporal filter's parameters
\STATE Initialize $M$ to zero
\STATE Order $E$ from the oldest to the newest event
\FOR  {$e$ in $E$}
    \FOR {$i$ in ($x_e-S$,$x_e+S$)}
        \FOR {$j$ in ($y_e-S$, $y_e+S$)}
            \IF { not ($i == x_e$ and $j==y_e$)}
                \STATE $M[i][j]= t_e$
            \ENDIF
        \ENDFOR
    \ENDFOR
    \IF {$t_e - M[x_e][y_e]>T$}
        \STATE Remove $e$ from $E$
    \ENDIF
\ENDFOR
\end{algorithmic}
\end{small}
\end{algorithm}
\vspace*{-15pt}
\end{figure}

\subsection{Adversarial Attacks in the Spatio-Temporal Domain}
\label{subsec:attack_video}

Currently, adversarial attacks are deployed on a wide range of deep learning applications. They represent a serious threat for safety-critical applications, like surveillance, medicine, and autonomous driving~\cite{Cheng2018SafetyCriticalDNN}\cite{RobustML_shafique}. The objective of a successful attack is to generate small perturbations to fool the network. Recently, adversarial attacks for SNNs have been explored. Bagheri et al.~\cite{Bagheri2018AdvTrainingSNN} and Marchisio et al.~\cite{Marchisio2019SNNAttack} analyzed adversarial attacks for SNNs in white-box and black-box settings, respectively. 
Sharmin et al.~\cite{Sharmin2019ACA} proposed a methodology to perform the adversarial attack on (non-spiking) DNNs, and then the DNN-to-SNN conversion made the adversarial examples craft the SNNs. 
Liang et al.~\cite{Liang2020ExploringAA} proposed a gradient-based adversarial attack methodology for SNNs. Venceslai et al.~\cite{Venceslai2020NeuroAttack} proposed a methodology to attack SNNs through bit-flips triggered by adversarial perturbations. \textit{However, none of these previous works analyze the attacks on frames of events, coming from DVS cameras.}

\begin{figure}[h]
    \includegraphics[width=\linewidth]{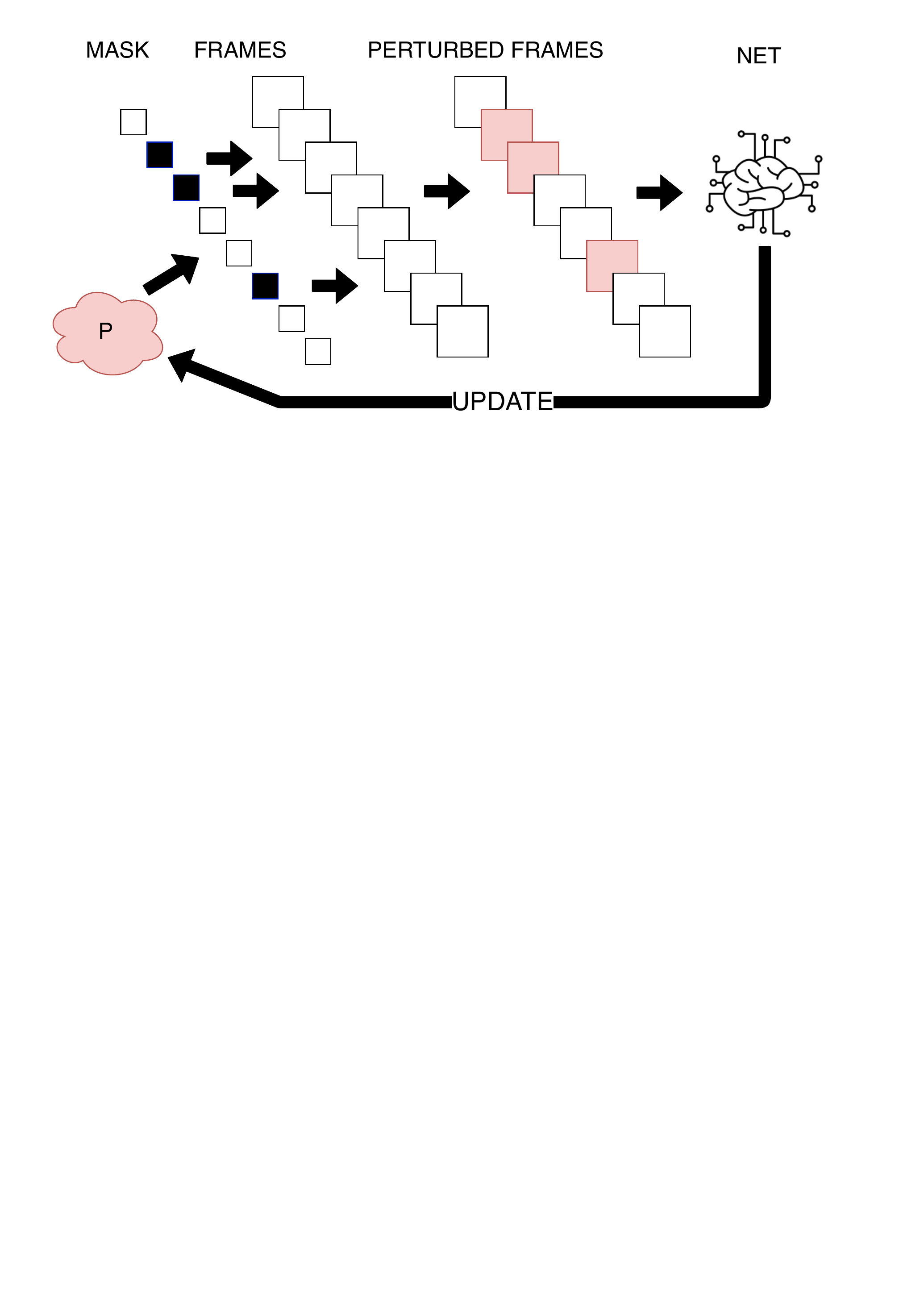}
    \vspace*{-245pt}
    \caption{Overview of the attack scheme for videos~\cite{Wei2019SparseAP}.}
    \label{fig:Attack}
    \vspace*{-10pt}
\end{figure}

For the adversarial attacks on images, the perturbations are introduced in the spatial domain only. However, when considering adversarial attacks on videos, which are sequences of frames, the attack algorithm is able to perturb in the temporal domain as well. While it is expected that the perturbations added to one frame propagate to other frames through temporal interaction, only perturbing a sparse subset of frames makes the attack stealthy. Indeed, state-of-the-art attacks on videos only add perturbations to a few frames, which are then propagated to other frames to misclassify the video~\cite{Wei2019SparseAP}. A simplified example, showing that a mask is generated in front of the frames for deciding which frames are perturbed and which not, is reported in Figure~\ref{fig:Attack}.


\section{R-SNN Methodology}

\subsection{Adversary Threat Models}
\label{subsec:threat_models}

\begin{figure*}[h]
    \centering
    \includegraphics[width=\linewidth]{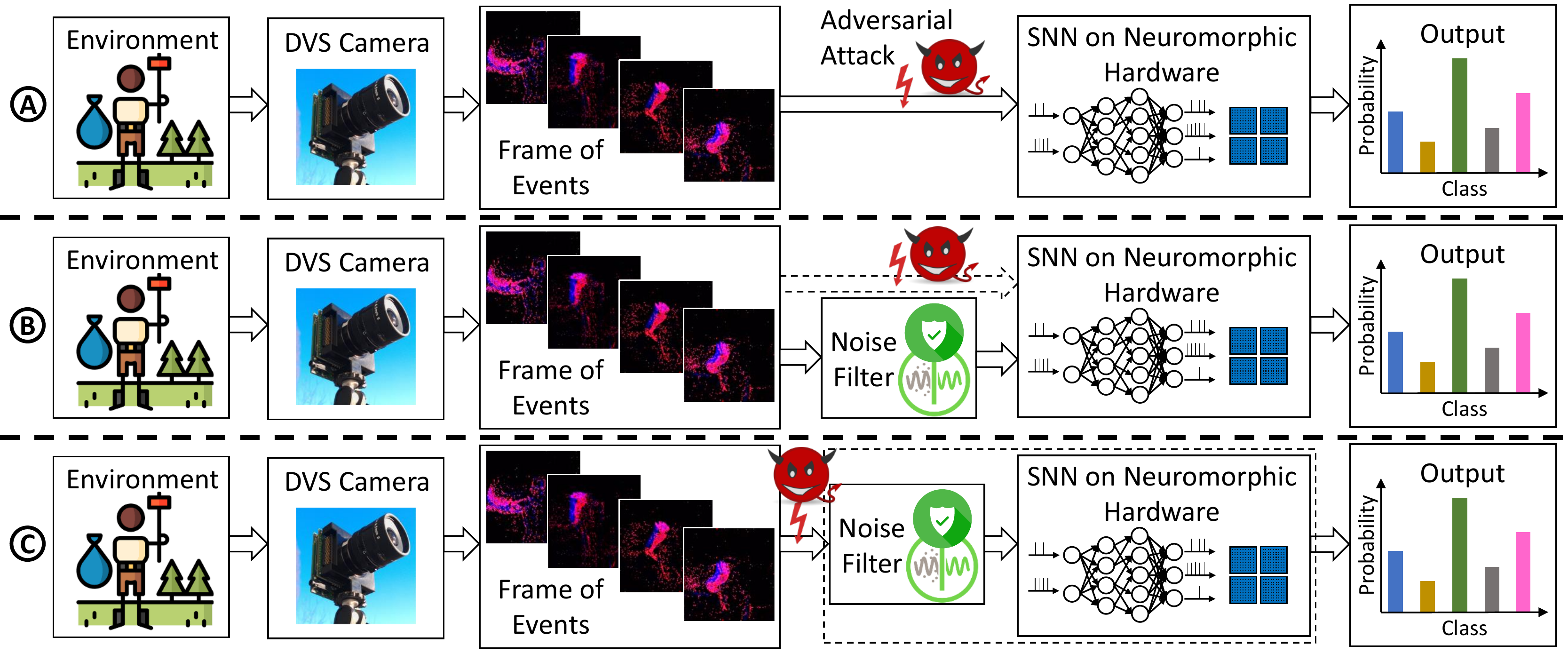}
    \caption{Adversarial threat models considered in this work. (a) The adversary introduces adversarial perturbations to the frames of events which are at the input of the SNN. (b) The noise filter is inserted as a defense to secure the SNNs against adversarial perturbations, while the adversary is unaware of the filter. (c) The adversary is aware of the presence of the noise filter, and sees it as a preprocessing step of the SNN.}
    \label{fig:threat_model}
    \vspace*{-10pt}
\end{figure*}

In our experiments, we assume different threat models in the system setting, which are shown in Fig.~\ref{fig:threat_model}. In all three scenarios, the given adversarial attack algorithm perturbs the frames of events generated from the DVS camera, with the aim of fooling the SNN. In the threat model~\rpoint{A}, the attacker has access to the frames of events at the input of the SNN. In the threat model~\rpoint{B}, the DVS noise filter is inserted in the system in parallel to the adversarial perturbation conducted by the attacker. It means that the attacker is unaware of the filter. Since under this assumptions the attack could be relatively weak, we analyze also the threat model~\rpoint{C}, in which the attacker is aware of the presence of the DVS noise filter. In such a scenario, the filter is seen as a preprocessing step of the SNN, and therefore is embedded in the attack loop.

\subsection{Adversarial Attack Generation for Frames of Events}
\label{subsec:attack_DVS}

The generation procedure for the adversarial attack for frames of events works as follows. Inspired by the algorithms of attacks for frame-based videos discussed in Section~\ref{subsec:attack_video}, we devise the specialized algorithm for the DVS signal. Algorithm~\ref{alg:AdvAttack} describes the step-by-step procedure of our methodology. It is an iterative algorithm, which progressively updates the perturbation values based on the loss function (lines~6-12), for each frame series of the dataset $D$. A mask $M$ determines in which subset of frames of events the perturbation should be added (line~7). Then, the output probability and the respective loss, obtained in the presence of the perturbation, are respectively computed in lines~9 and~10. Finally, the perturbation values are updated based on the gradients of the inputs with respect to the loss.

\begin{figure}[h]
\vspace*{-4pt}
\begin{algorithm}[H]
\caption{\textbf{:} The SNN Adversarial Attack Methodology.}
\label{alg:AdvAttack}
\begin{small}
\begin{algorithmic}[1]
\STATE Being $M$ a mask able to select only certain frames
\STATE Being $D$ a dataset composed of DVS images
\STATE Being $P$ a perturbation to be added to the images
\STATE Being $prob$ the output probability of a certain class
\FOR  {$d$ in $D$}
    \FOR {$i$ in $max\_iteration$}
    \STATE Add $P$ to $d$ only on the frames selected by $M$
    \STATE Calculate the prevision on the perturbed input
    \STATE Extract $prob$ for the actual class of $d$
    \STATE Update the loss value as $loss=-log (1- prob$)
    \STATE Calculate the gradients and update $P$
    \ENDFOR
\ENDFOR
\end{algorithmic}
\end{small}
\end{algorithm}
\vspace*{-15pt}
\end{figure}

\subsection{Our Proposed Defense Methodology}
\label{subsec:methodology}

Our methodology for defending SNNs is based on specialized DVS-noise filtering. The details for selecting efficient values of the spatial parameter $s$ and temporal parameter $t$ of the filter are reported in Algorithm~\ref{alg:defense_method}. For different threat models, it automatically searches for the best combination of $s$ and $t$, by applying the attack in the presence of the filter with the given parameters. The accuracy of the SNN in such conditions is compared to the previously-recorded highest accuracy (line~14 of Algorithm~\ref{alg:defense_method}). At the output, the parameters $s'$ and $t'$ which provide the highest accuracy are found.

\begin{figure}[h]
\vspace*{-4pt}
\begin{algorithm}[H]
\begin{small}
\begin{algorithmic}[1]
\STATE Being $M$ the collection of adversarial threat models
\STATE Being $A$ the adversarial attack
\STATE Being $F(s,t)$ a DVS noise filter with spatial parameter $s$ and temporal parameter $t$
\STATE Being $\mathcal{S}$ the set of possible values of $s$
\STATE Being $\mathcal{T}$ the set of possible values of $t$
\STATE Being $N(F)$ the SNN that we want to robustify with $F$
\FOR{$m$ in $M$}
    \STATE Set the relative positions of $A$ and $F$, based on $m$
    \STATE $Acc'=0$
    \STATE $s'=0$
    \STATE $t'=0$
    \FOR{$s$ in $\mathcal{S}$}
        \FOR{$t$ in $\mathcal{T}$}
            \IF{Accuracy$(N(F(s,t))) \geq Acc'$}
                \STATE $Acc'=$Accuracy$(N(F(s,t)))$
                \STATE $s'=s$
                \STATE $t'=t$
            \ENDIF
        \ENDFOR
    \ENDFOR
    \STATE \textbf{Output:} Values $s'$ and $t'$ for a robust defense in $m$
\ENDFOR
\end{algorithmic}
\end{small}
\caption{: The SNN Defense Methodology.}
\label{alg:defense_method}
\end{algorithm}
\vspace*{-15pt}
\end{figure}

\section{Evaluation of the R-SNN methodology}
\label{sec:results}


\subsection{Experimental Setup}

\begin{figure}[h]
    \vspace*{5pt}
    \includegraphics[width=\linewidth]{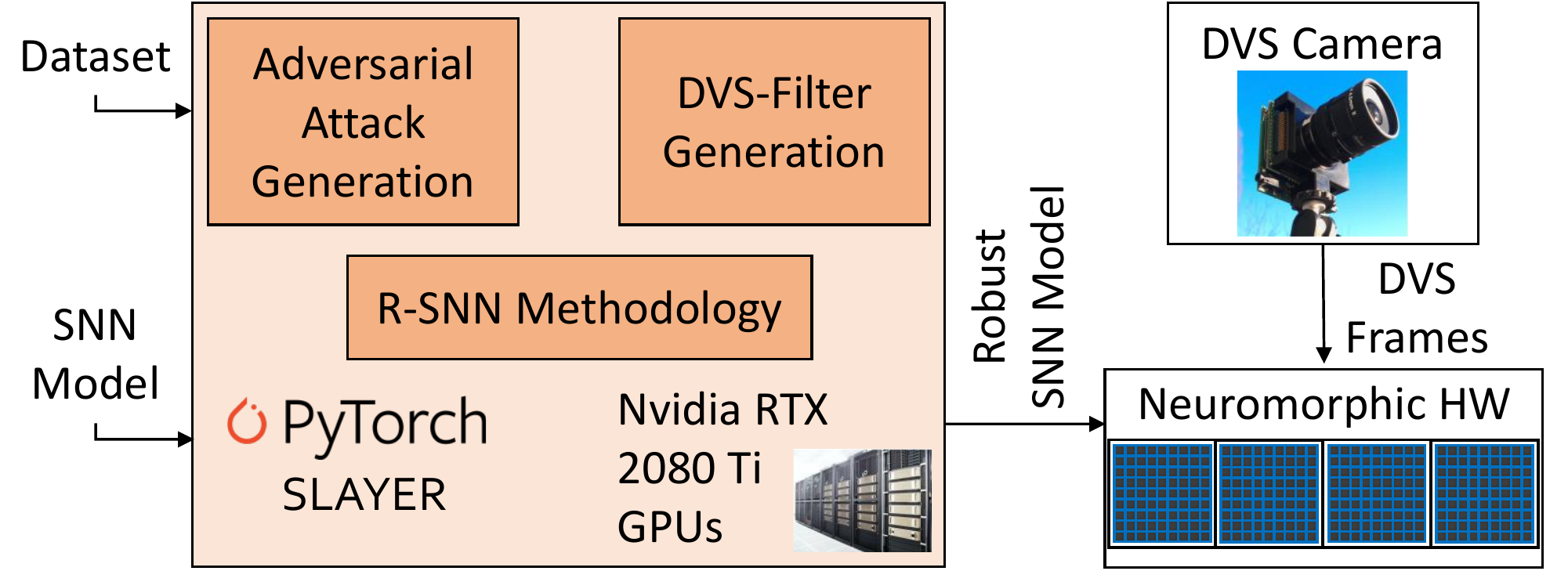}
    \caption{Experimental setup, tool-flow, and integration with the system.}
    \label{fig:exp_setup}
    \vspace*{-10pt}
\end{figure}

In our experiments, we used two event-based dataset, the DVS-Gesture~\cite{Amir2017DVSgesture} and the NMNIST~\cite{Orchard2015NMNIST}. The former is a collection of of 1077 samples for training and 264 for testing, divided into 11 classes, while the latter is a spiking version of the original frame-based MNIST dataset~\cite{LeCun1998MNIST}. It contains 60,000 training and 10,000 testing samples generated by an ATIS event-based sensor~\cite{Posch2011ATIS} that is moved while capturing the MNIST images projected on a LCD screen. 
For the DVS-Gesture dataset, we considered the 4-layer SNN as described in~\cite{Shrestha2018SLAYER}, with two convolutional layers and two fully-connected layers. It has been trained it for 625 epochs with the SLAYER backpropagation method~\cite{Shrestha2018SLAYER}, using a batch size of 4 and learning rate equal to 0.01. For the NMNIST dataset, we employed a spiking multilayer perceptron with two fully-connected layers~\cite{Shrestha2018SLAYER}, trained for 350 epochs with the SLAYER backpropagation method~\cite{Shrestha2018SLAYER}, using a batch size of 4 and learning rate equal to 0.01. 
We implemented the SNNs on a ML-workstation with two Nvidia GeForce RTX 2080 Ti GPUs, using the PyTorch framework~\cite{pytorch}. We also implemented the adversarial attack algorithm and the noise filter of~\cite{LinaresBarranco2019FilterDVS} in PyTorch. The experimental setup and tool-flow in a real-world setting is shown in Fig.~\ref{fig:exp_setup}.

\subsection{SNN Robustness under Attack Without the Noise Filter}
For the threat model \rpoint{A}, the attacker introduces the adversarial perturbations directly to the input of the SNN. In this case, the SNN for the DVS-Gesture dataset is not protected by the filter and the accuracy dropped to $15.15\%$ (see pointer~\rpoint{1} in Fig.~\ref{fig:accuracy}a). A similar behavior is noted on the SNN for the NMNIST dataset, where the attack reduces the accuracy to 4\% (91\% reduction, as highlighted by pointer~\rpoint{6} in Fig.~\ref{fig:accuracy}b). We noticed that for both datasets the largest accuracy drop is obtained already after the first iteration of the attack algorithm. Further iterations of the algorithm do not appear to reduce the accuracy to a greater extent.

\subsection{SNN Robustness under Attack by Noise Filter-Unaware Adversary}
\begin{figure}[h]
    \includegraphics[width=\linewidth]{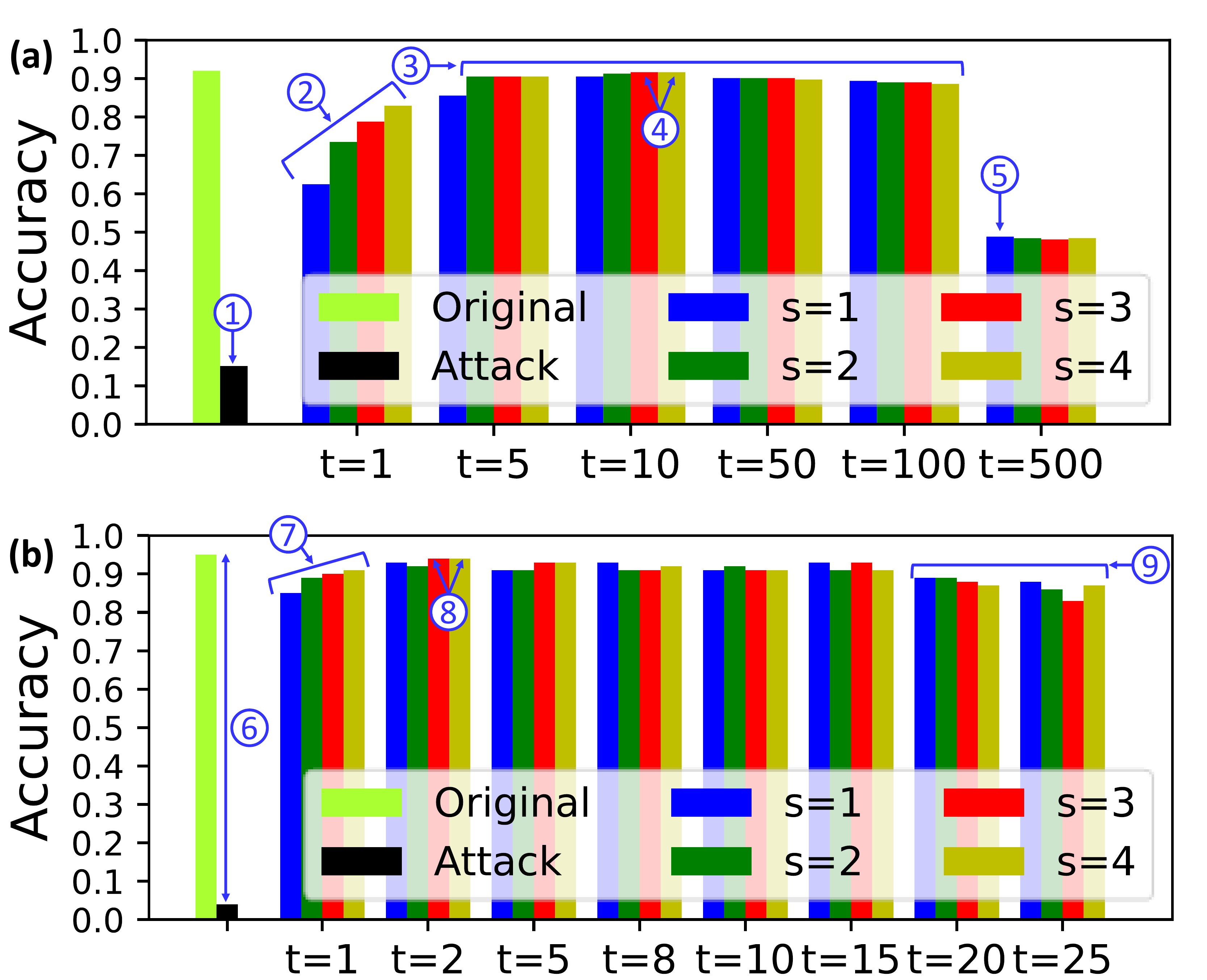}
    \caption{SNN robustness under the adversarial threat model A, and under the threat model B with different parameters $s$ and $t$ of the filter. (a) Results for the DVS-Gesture dataset. (b) Results for the NMNIST dataset.}
    \label{fig:accuracy}
    \vspace*{-10pt}
\end{figure}

Afterward, we analyzed the SNN robustness for the threat model~\rpoint{B}, that is the case in which the attacker is able to introduce a perturbation on the input, but is not aware of the presence of the DVS filter. For this experiment set, the accuracy was much higher than for the threat model~\rpoint{A}, proving the effectiveness of the filter as a defense method, for guaranteeing a high robustness of the SNN. The results obtained with our proposed \textit{R-SNN} methodology, varying both the parameters $s$ and $t$ of the filters, are reported in Fig.~\ref{fig:accuracy}. 
On the SNN for the DVS-Gesture dataset, for a wide variety of values of $s$ and $t$ (see pointer~\rpoint{3}), the accuracy does not change much, settling around  $90\%$, while with $t=500$ it dropped to $48\%$ (see pointer~\rpoint{5}).
However, when $t=1$ the influence of $s$ is more evident (see pointer~\rpoint{2}). In fact, the accuracy scales from $62.5\%$ when $s=1$ to $83\%$ when $s=4$. In all the other cases, the difference is almost not noticeable. Notice, though, that the higher $s$ is, the slower the filter is to process all the data. Among the considered values, $t = 10$ produced the highest accuracy for every $s$, peaking at $91.67\%$ with $s=3$ and $s=4$ (see pointer~\rpoint{4}). 
On the SNN for the NMNIST dataset, a similar behavior is shown. For $t=1$, the accuracy strongly depends on $s$ (see pointer~\rpoint{7}). The peak of 94\% accuracy is reached for $(s,t)=(3,2)$ and $(s,t)=(4,2)$ (see pointer~\rpoint{8}). Note that, this is only 1\% lower than the original accuracy, i.e., with clean inputs. On the other hand, the accuracy drops below 90\% for $t \geq 20$ (see pointer~\rpoint{9}).

\subsection{SNN Robustness under Attack by Noise Filter-Aware Adversary}

\begin{figure}[h]
    \includegraphics[width=\linewidth]{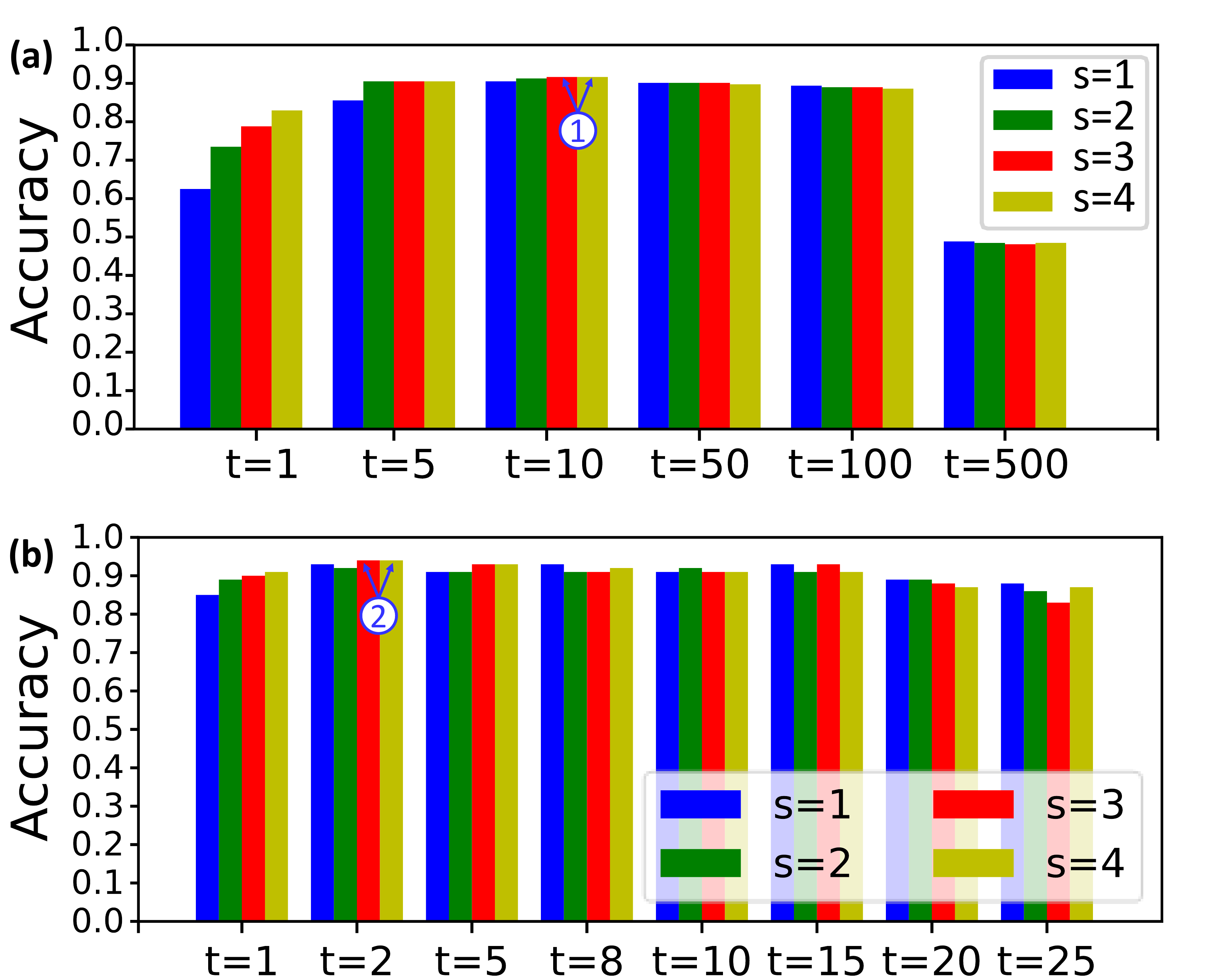}
    \caption{SNN robustness under adversarial threat model C. (a) Results for the DVS-Gesture dataset. (b) Results for the NMNIST dataset.}
    \label{fig:accuracy_1}
    \vspace*{-10pt}
\end{figure}

We also evaluated the \textit{R-SNN} methodology on the threat model~\rpoint{C}, in which the attacker is aware of the presence of the filter. This time the filter was seen as an integral part of the SNN, more specifically as a preprocessing stage. As expected, also in this scenario the filter is effective as a defense mechanism. The differences w.r.t. the threat model~\rpoint{B} are not noticeable. Among the experiments for the DVS-Gesture dataset, the highest robustness is reached for $(s,t)=(3,10)$ and $(s,t)=(4,10)$, where the SNN exhibits an accuracy of $91.67\%$ (see pointer~\rpoint{1} in Fig.~\ref{fig:accuracy_1}a). 
For the NMNIST dataset, the highest robustness, i.e., with an accuracy of 94\%, is measured for $(s,t)=(3,2)$ and $(s,t)=(4,2)$ (see pointer~\rpoint{2} in Fig.\ref{fig:accuracy_1}b).
Such a result is a clear sign that this kind of attack is not able to overcome the presence of the filter. Therefore, the attack algorithm is not able to effectively learn the filter's functionality through a gradient-based approach, even though being aware of it. 

\begin{figure*}[t]
    \centering
    \includegraphics[width=\linewidth]{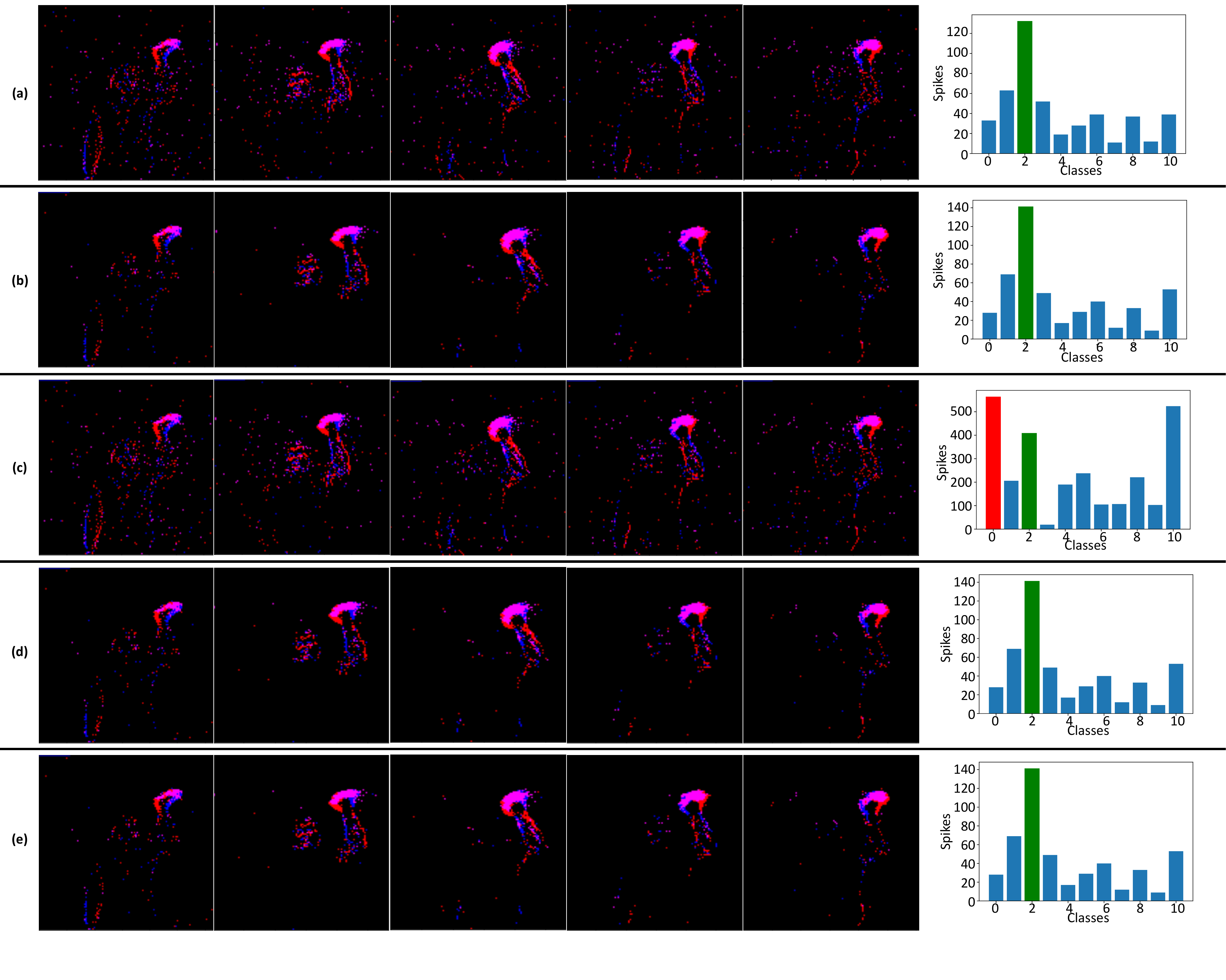}
    \vspace*{-30pt}
    \caption{Detailed example of a sequence of event labeled as \textit{left hand wave}. On the left, the frames of events are shown. The histograms on the right-most column report the number of spikes emitted by the neurons of the last layer, which correspond to the output classes. (a) Clean event series. (b) Event series filtered with $s=2$ and $t=5$. (c) Event series under the adversarial threat model A, unfiltered. (d) Event series under the adversarial threat model B, filtered with $s=2$ and $t=5$. (e) Event series under the adversarial threat model C, filtered with $s=2$ and $t=5$.}
    \label{fig:example_left_hand_wave}
    \vspace*{-0pt}
\end{figure*}

\subsection{Case Study: Output Probability Variation}

To investigate more in details the effect of the adversarial attack and the filter, we show a comprehensive case study on a test DVS-gesture sample labeled as \textit{left hand wave}. Fig.~\ref{fig:example_left_hand_wave} reports the frames of events and output probabilities for each adversarial threat model presented in this paper, as well as for the clean inputs and the filtered event series without attack. For the clean images the SNN correctly classifies the events as the class 2, which corresponds to \textit{left hand wave} (see Fig.~\ref{fig:example_left_hand_wave}-a). By filtering the input signal with $s=2$ and $t=5$, as shown in \mbox{Fig.~\ref{fig:example_left_hand_wave}-b}, the frames of events are visibly different than the previous case. However, the changes in the output probabilities is minimal, and therefore the SNN correctly classifies the input. 
When the attack is applied, the output probability of the class 0, which corresponds to \textit{hand clap}, overcomes the correct class. Note that, despite a great difference in the output probabilities, the modifications of the frames of events, compared to the clean event series, are barely noticeable (see Fig.~\ref{fig:example_left_hand_wave}-c). However, in the presence of the filter under the adversarial threat models~\rpoint{B} and~\rpoint{C}, the SNN correctly classifies the input. The high gap in the probabilities between the correct class and the other classes in Figures~\ref{fig:example_left_hand_wave}-d and~\ref{fig:example_left_hand_wave}-e is an indicator for the high robustness of our defense method.

\section{Conclusion}

In this paper, we presented \textit{R-SNN}, a defense methodology for Spiking Neural Network (SNN) based systems using the event-based Dynamic Vision Sensors (DVS). The proposed gradient-based adversarial attack algorithm exploits the spatio-temporal information residing in the DVS signal, and mislead the SNN, while generating small imperceptible differences w.r.t. the clean series of events. The \textit{R-SNN} defense is based on specialized DVS-noise filters, and an automatic selection of the filter parameters lead to high SNN robustness against adversarial attacks, under different threat models and different datasets. These findings consolidate the positioning of SNNs as robust and energy-efficient solutions, and might enable more advanced secure SNN designs. We release the source code of the \textit{R-SNN} methodology at \url{https://github.com/albertomarchisio/R-SNN}.

\section*{Acknowledgments}

This work has been partially supported by the Doctoral College Resilient Embedded Systems, which is run jointly by the TU Wien's Faculty of Informatics and the UAS Technikum Wien. This work was also jointly supported by the NYUAD Center for Interacting Urban Networks
(CITIES), funded by Tamkeen under the NYUAD Research Institute Award CG001 and by the Swiss Re Institute under the Quantum Cities™ initiative, and Center for CyberSecurity (CCS), funded by Tamkeen under the NYUAD Research Institute Award G1104.

\begin{refsize}
\bibliographystyle{ieeetr}
\bibliography{main.bib}
\end{refsize}

\end{document}